\documentclass{artificial-life}
\usepackage[utf8]{inputenc}
\DeclareUnicodeCharacter{0300}{\`{}}  % Unicode U+0300 = combining grave
\usepackage[style=apa,natbib=true]{biblatex}
\usepackage{color}
\usepackage{amsmath,amsthm,ascmac,amssymb}

\usepackage{algorithm}
\usepackage{algpseudocode}
\usepackage[protect,final]{changes} %おわったら final
\setaddedmarkup{\textcolor{red}{#1}}

\title{System 0/1/2/3: Quad-process theory \\for multi-timescale embodied collective cognitive systems}

\auth{
Tadahiro Taniguchi \affil{1,2} \twitter{tanichu}, %
Yasushi Hirai \affil{3}, %
Masahiro Suzuki \affil{4}, %
Shingo Murata \affil{5}, %
Takato Horii \affil{4,6}, %
Kazutoshi Tanaka \affil{7}
}

\corresponding{Tadahiro Taniguchi}{taniguchi@i.kyoto-u.ac.jp}

\affiliations{
    \item Kyoto University, Yoshida-honmachi, Sakyo-ku, Kyoto, Kyoto, Japan
    \item Ritsumeikan University, 1-1-1 Noji-higashi, Kusatsu, Shiga, Japan
    \item Keio University, 2-15-45 Mita, Minato-ku, Tokyo, Japan
    \item The University of Tokyo, 7-3-1 Hongo, Bunkyo-ku, Tokyo, Japan
    \item Keio University, 3-14-1 Hiyoshi, Kohoku-ku, Yokohama, Kanagawa, Japan
    \item Osaka University, 1-3 Machikaneyama, Toyonaka, Osaka, Japan
    \item OMRON SINIC X Corporation, 5-24-5 Hongo, Bunkyo-ku, Tokyo, Japan
}

\abs{
This paper introduces the System 0/1/2/3 framework as an extension of dual-process theory, employing a quad-process model of cognition. Expanding upon System 1 (fast, intuitive thinking) and System 2 (slow, deliberative thinking), we incorporate System 0, which represents pre-cognitive embodied processes, and System 3, which encompasses collective intelligence and symbol emergence. We contextualize this model within Bergson's philosophy by adopting multi-scale time theory to unify the diverse temporal dynamics of cognition.  

System 0 emphasizes morphological computation and passive dynamics, illustrating how physical embodiment enables adaptive behavior without explicit neural processing. Systems 1 and 2 are explained from a constructive perspective, incorporating neurodynamical and AI viewpoints. In System 3, we introduce collective predictive coding to explain how societal-level adaptation and symbol emergence operate over extended timescales.  

This comprehensive framework ranges from rapid embodied reactions to slow-evolving collective intelligence, offering a unified perspective on cognition across multiple timescales, levels of abstraction, and forms of human intelligence. The System 0/1/2/3 model provides a novel theoretical foundation for understanding the interplay between adaptive and cognitive processes, thereby opening new avenues for research in cognitive science, AI, robotics, and collective intelligence.
}


\begin{document}

% bring the following back befor submission
% Don't Touch ! %%%%%% 
\coverpage           % %
%\linenumbers         % 
%\doublespacing       % 
%%%%%%%%%%%%%%%%%%%%%% 

%%% ALL TEXT OF ARTICLE BELOW THIS LINE %%%
%-----------------------------------------%

\section{Introduction}
To understanding intelligence—whether biological or artificial—the diverse timescales at which cognition and adaptation occur must be considered. Existing models of cognition, such as dual-process theory, have significantly advanced our understanding of human intelligence by distinguishing between fast intuitive thinking (System 1) and slow deliberative reasoning (System 2)~\citep{kahneman2011thinking,bengio2019system2}. However, these models remain limited in scope:
\begin{itemize}
    \item They primarily focus on internal cognitive processes, overlooking embodiment and sensorimotor interactions with the environment.
    \item They do not account for collective intelligence, in which language and symbol systems evolve dynamically within human societies.
    \item They lack a broader temporal hierarchy, failing to capture how cognitive processes unfold across multiple timescales, from rapid reflexive reactions to the slow evolution of societal knowledge.
\end{itemize}
These limitations become more pronounced as we aim to develop AI and robotics capable of better real-world adaptation. Embodied AI systems, such as robots, have difficulty with adaptive real-world behavior because their cognitive models often fail to integrate fast, reactive motor control with slow, deliberative planning. Moreover, AI and robots face challenges in modifying the languages they use for communication and in contributing to the cultural and societal formation necessary for collective survival. Their cognitive models often fail to manage long-term collective semiotic dynamics—i.e., symbol emergence—hindering their ability to flexibly adapt language and symbolic systems to their environment.

To address these challenges, we propose the System 0/1/2/3 framework, which is a quad-process model of cognition that extends dual-process theory by integrating pre-cognitive embodied processes (System 0) and collective intelligence and symbol emergence (System 3). This framework is inspired by Bergson''s philosophy of time, which provides a theoretical foundation for understanding how cognition operates at multiple temporal scales, from super-fast embodied reactions to super-slow collective intelligence (see Section~\ref{sec:mts}).

\deleted{Inside the brain, the free-energy principle proposed by Karl Friston provides a unifying framework for understanding adaptive behavior in biological systems. }
\deleted{This principle suggests that intelligent agents strive to minimize surprise by continuously updating their internal world models. }
\deleted{These world models are essential in both biological and artificial intelligence, enabling prediction, planning, and adaptive behavior.}

\added{Inside the brain, the free-energy principle (FEP) introduced by Karl Friston offers a unifying account of how biological systems remain adaptive~\citep{friston2010free,Parr2022}. According to FEP, an agent minimizes expected surprise by continually adjusting the parameters of a generative process that couples the organism to its niche. Whether one reads this process as an “internal model” or as a non-representational, action-oriented dynamic is left open—FEP itself is neutral on that point~\citep{Constant2021}. What matters is that this temporally deep generative process supports prediction, planning, and other forms of adaptive behavior in both living organisms and artificial systems~\citep{friston2021world,taniguchi2023world}.}

However, cognition is not limited to the processes within an individual brain. Humans achieve long-term coordination and adaptation at the collective level by communicating, establishing rules, naming things, and creating agreements. These processes often rely on symbols, typically represented by language. In semiotics, it is widely recognized that symbol systems have \emph{arbitrariness}, meaning that the relationship between the signifier and signified is not fixed and evolves across communities and time periods~\citep{Chandler2002}.
We emphasize this \emph{arbitrariness} to highlight the core argument of this paper: the existence of super-slow dynamics in human (collective) intelligence. Human societies continually update shared symbol systems, i.e., systems of signs and their relationships with phenomena, over time through social interactions. This arbitrariness allows adaptive flexibility in semiotic communication. Humans utilize this semiotic plasticity, analogous to neural plasticity in the brain, to engage in intelligent activities that no single individual can achieve alone. We refer to this level as System 3 later in this paper.

Additionally, interactions between the brain, body, and environment demonstrate self-organizational and adaptive behavior, even without neural learning. For example, passive dynamic walkers can descend slopes without requiring computational control~\citep{mcgeer1990passive,shintake2018soft}. These examples illustrate how body-environment interactions contribute to flexible, robust, and intelligent behavior generation. Although foundation models and multimodal LLMs are gaining attention~\citep{bommasani2021opportunities,radford2021learning}, robotics foundation models primarily focus on computational data processing and neglect embodied interactions~\citep{driess2023palm,arai2024covla,kim2024openvla,dey2024revla,zhen20243d,Kawaharazuka16092024}. However, physical interactions that do not require neural processing exhibit super-fast dynamics. Recent AI studies have often overlooked this layer, which we refer to as System 0 in this paper.

The newly introduced terms of Systems 3 and 0 contrast with Systems 1 and 2 from dual-process theory~\citep{kahneman2011thinking}. By considering self-organizational dynamics both inside and outside the brain, we move beyond the computational view of intelligence to a self-organizational perspective.

Furthermore, by connecting this richer hierarchical multi-timescale understanding of cognitive systems to Bergson''s philosophy, we aim to establish a reciprocal dialogue between Bergson''s theories of memory, the emergence of consciousness, and discussions on cognitive systems and intelligence. Additionally, we attempt to extend Bergson''s multi-timescale framework by examining both its similarities to and differences from contemporary cognitive models.

\added{While the concept of System 3 is rooted in the Collective Predictive Coding (CPC) hypothesis introduced in Taniguchi (2024), this paper's primary contribution is to extend the traditional System 1/2 framework into a comprehensive quad-process model. The key novelties are: first, the introduction of System 0 to account for pre-cognitive embodied processes, thereby completing the 0/1/2/3 architecture; and second, the grounding of this entire multi-timescale framework in an extended interpretation of Bergson's philosophy of time, as detailed in~\ref{sec:mts}.}

\deleted{This paper extends the traditional System 1/2 framework by incorporating Systems 0 and 3 to clarify the challenges in AI, robotics, and collective intelligence. This provides new insights into Bergson's philosophy of time, which we discuss in Section~\ref{sec:mts}.}

The main contributions of this paper are two-fold:
\begin{itemize}
    \item This paper extends the traditional dual-process theory by introducing the \emph{System 0/1/2/3} framework, a quad-process model of cognition. Building upon the well-established \emph{System 1} (fast, intuitive thinking) and \emph{System 2} (slow, deliberative thinking), we incorporate \emph{System 0}, which represents pre-cognitive embodied processes occurring at super-fast timescales, and \emph{System 3}, which encompasses collective intelligence and symbol emergence operating at super-slow timescales. This framework provides a comprehensive model for understanding cognition across multiple temporal scales, from rapid sensorimotor interactions to the slow evolution of collective knowledge.
    
    \item This paper further extends \emph{Bergson''s multi-timescale (MTS) interpretation} to the System 0/1/2/3 framework, offering a philosophically grounded perspective on cognition that unifies \emph{individual, embodied, and collective intelligence}. By integrating Bergson''s view of intrinsic timescales with modern cognitive science, AI, and artificial life, we provide a novel perspective on how cognitive processes operate across distinct but interconnected temporal layers. This synthesis elucidates the role of \emph{morphological computation and passive dynamics} in System 0, \emph{predictive and deliberative cognition} in Systems 1 and 2, and \emph{symbol emergence and collective predictive coding (CPC)} in System 3. This theoretical foundation offers a new lens for understanding cognition as an emergent, multi-scale phenomenon, bridging the gap between embodied intelligence, computational neuroscience, and society-level adaptation.
\end{itemize}

The remainder of this paper is organized as follows.  
Section~\ref{sec:bg} provides a background and related work, reviewing existing artificial cognitive models related to dual-process theory, embodied cognition, collective intelligence involving emergent language, and Bergson's philosophy.  
Section~\ref{sec:system0123} introduces the System 0/1/2/3 framework, presenting its theoretical foundation and explaining how it extends dual-process theory to incorporate embodied and collective intelligence.  
Section~\ref{sec:mts} explores the extended Bergsonian MTS interpretation, providing a philosophical foundation for understanding how cognitive processes operate across different temporal scales.  
Section~\ref{sec:discussion} discusses the implications and challenges of this framework, particularly in the contexts of AI, robotics, and artificial life.  
Finally, Section~\ref{sec:conclusion} concludes the paper and outlines potential future research directions.

\section{Background and Related Work}\label{sec:bg}

\subsection{Dual-process theory}\label{sec:system12}
Dual-process theory has been widely discussed across various domains, including cognitive science and behavioral economics~\citep{sloman1996empirical,evans2008dual,evans2013dual,kahneman2011thinking}. In the context of developing artificial cognitive systems that incorporate dual-process mechanisms, this section provides an overview of representative studies and discussions related to dual-process theory in AI and cognitive robotics.
\subsubsection{Fast and Slow in AI and Cognitive systems}\label{sec:system12-ai}
Early AI systems were symbolic, based on logical reasoning and search algorithms, corresponding to ``slow'' processes that involved time-consuming search and analysis. Later, with the development of machine and deep learning, ``fast'' reasoning became possible in fields such as image recognition and natural language processing.
However, despite recent advances in deep learning, challenges such as generalization to unseen and out-of-distribution environments and causal reasoning remain unresolved when compared to human capabilities.

It has been argued that a combination of logical reasoning and deep learning is required to address these limitations.
\citet{bengio2017consciousness} suggested the need to generalize from data to a consciousness stream consisting of a small number of sparsely connected concepts. Hybrid systems that integrate machine learning with symbolic reasoning, known as neuro-symbolic AI, have been actively investigated ~\citep{sheth2023neurosymbolic}. 
Symbolic AI is considered to correspond to System 2 and machine learning to System 1~\citep{sheth2023neurosymbolic}. Systems 1 and 2 are sometimes considered data-driven and knowledge-driven, respectively.

Several frameworks have been proposed to implement System 2 using System 1. One of the most straightforward approaches is to use deep learning to acquire representations from the external world and then process these representations using symbolic AI. For symbolic AI to effectively manipulate these representations, it is necessary to acquire independent representations of each factor, a concept known in deep learning as disentangled representations~\citep{bengio2013representation}. Disentangled representations allow the system to clearly distinguish between different factors, making reasoning and manipulation more straightforward. For example, \citet{higgins2017beta} demonstrated that disentangled representations in generative models, such as the $\beta$-VAE, enable a clear separation of independent factors of variation in data, thereby improving the ability to perform tasks such as causal reasoning and transfer learning. \citet{locatello2019challenging} highlighted that although fully unsupervised disentanglement is challenging, leveraging some form of inductive bias or weak supervision can significantly improve the learning of disentangled representations.

However, simply abstracting information from the external world may be insufficient. In practice, for an agent to adapt to a new environment, it must not only abstract the knowledge it has learned but also instantiate that knowledge within the new environment~\citep{booch2021thinking}. This requires the abilities to recognize similarities and differences between two environments and to determine what aspects should be forgotten during abstraction.
Bengio's ``consciousness prior theory'' suggests that such representation learning can be achieved by using conscious processing as a prior, which can be implemented through attention mechanisms~\citep{bengio2017consciousness}. 
Recurrent independent mechanisms (RIMs) introduce multiple recurrent neural network (RNN)-based modules that are sparsely connected based on attention mechanisms~\citep{madan2021fast}. These modules operate independently and are activated only when relevant to the task, allowing for more efficient and specialized processing.
In addition, a framework that generalizes abstracted knowledge as an autonomous skill that can be applied in new environments is necessary.\citet{ellis2021dreamcoder} proposed a method where programming languages are viewed as primitive skills and neural networks are used to guide the abstraction process. They demonstrated that this approach enables the automatic acquisition of skills applicable to new environments.

In recent years, there has been active research aiming to relate large language models (LLMs) to the discussion of System 1 and System 2. Generally, because LLMs primarily generate tokens probabilistically (i.e., predict the next word) based on contextual cues, they are often seen as mimicking the intuitive, automatic processes of System 1. However, recent studies suggest that LLMs are also capable of carrying out System 2 tasks involving more step-by-step reasoning and deliberation. 

% LLMにおける段階的な思考
To enable System 2-like reasoning, approaches such as chain-of-thought and tree-of-thought have been proposed~\citep{wei2022chain, yao2024tree}. Rather than immediately providing an answer to a question all at once, these methods introduce a mechanism that explicitly outlines intermediate reasoning steps and verifies them incrementally until a conclusion is reached. Similar to how humans logically develop arguments in their minds, the model advances through multiple reasoning stages, making contradictions less likely to arise and thus making these approaches particularly suitable for tackling difficult mathematical problems as well as tasks involving long-text comprehension and summarization. Notably, simulating a tree-of-thought within prompts has been reported to enable LLMs to exhibit a more structured and deliberative inference process~\citep{long2023large}.

% LLMにおける自己反省
Self-refine or self-reflection, first introduced as ``Self-Refine'' by \citep{madaan2023self}, in LLMs can also be considered a method for enabling System 2-type reasoning. In this approach, the model itself examines whether its own output is correct or contains contradictions, and revises it as needed. Because the model first generates an answer and then reviews it, the process not only relies on intuition (System 1) but also incorporates deliberation and re-evaluation characteristic of System 2. A notable follow-up is Reflexion~\citep{shinn2023reflexion}, which extends self-refine by letting LLMs revise their outputs multiple times. After each attempt, the model stores any mistakes or uncertainties and consults this record in future steps, thereby improving subsequent responses through repeated self-examination.

% 外部ツールを使う話（意識的・段階的処理という意味で関連）
In methods such as ReAct~\citep{yao2023react} and ToolFormer~\citep{schick2023toolformer}, LLMs acquire additional information or call external computational tools (e.g., Python, calculators, search engines) to improve their output. This is similar to human System 2 in that it implements a conscious, step-by-step process of not just generating answers intuitively (System 1), but also pausing to think about what information is lacking, gathering information from outside sources, and re-evaluating. ReAct is a prompting method that interleaves a LLM’s reasoning process with external tool usage, thus integrating the two. By combining introspective reasoning with external tool utilization, the model can actively think at each step while retrieving the necessary information, enabling it to solve more complex problems than would be possible with chain-of-thought alone. Toolformer is a model that, through self-supervised learning, acquires the ability to autonomously decide which external API tool to use and when to call upon it. This can be likened to how humans, when facing computations they cannot handle entirely in their head, may resort to writing on paper or consulting a dictionary. By allowing a pause in the continuous stream of text generation to explicitly call an API and wait for its results, the model incorporates a step-by-step, planful thinking process.

% AIエージェントについて．長くなるのでいらないかも
Furthermore, the integration of reasoning with external tool usage has been actively advancing in the development of LLM agents in recent years~\citep{wang2024survey}. For example, LangChain~\citep{langchain} serves as a framework for integrating LLMs with external tools, incorporating prompt designs and tool interactions inspired by approaches like ReAct. Additionally, efforts such as Auto-GPT~\citep{autogpt2023} and BabyAGI~\citep{babyagi2023} have emerged to develop agents that engage in continuous reasoning and planning while interacting with external environments to achieve long-term goals. Unlike single-turn Q\&A models, these agents dynamically formulate plans, retrieve external resources as needed, and iteratively execute tasks to accomplish user-defined objectives. As a result, the "System 2"-like approach has become increasingly significant in the design and implementation of LLM agents.

% 以前の内容．現状の文に繋がらないのでいらない．
%Despite the general view that LLMs primarily engage in System 1 processing, some studies have argued that these models are also capable of performing System 2 tasks, which require deeper and more reflective thinking~\citep{hagendorff2023thinking}. In support of this, \citet{hulbert2023} proposed using prompts to simulate tree-of-thought reasoning, enabling LLMs to engage in more organized and deliberative processes. Furthermore, there are ongoing efforts to ``distill'' the outcomes of System 2-like thinking into System 1, which would allow for smoother transitions between deliberate reasoning and more intuitive decision-making.

\subsubsection{Fast and Slow in Neuro-dynamical Systems}\label{sec:system12-neuro}

In the field of cognitive neurorobotics, Tani et al. developed various neuro-dynamical systems based on RNNs~\citep{Tani2016}.
While their work primarily focused on the top-down intentional and bottom-up recognition processes underlying complex behavior and cognition in the brain, they also provided insights into how different levels of neural dynamics correspond to Systems 1 and 2.

Their approaches were inspired by dynamical systems theory for autonomous agent behavior (both artificial and biological), emphasizing the tight coupling between an agent's internal dynamics and the external dynamics of its environment~\citep{Beer1995, Beer1996}.
This original dynamical systems approach, which may correspond to Systems 0 and 1, explains an agent's lower-level or reflexive behavior in response to environmental changes through the concept of entrainment into attractors.
By combining this approach with the idea of forward models~\citep{Wolpert1995}, which correspond with the so-called world models used in recent model-based reinforcement learning frameworks~\citep{Ha2018b, Hafner2018, Hafner2020a}, early work on robotic navigation problems~\citep{Tani1996} demonstrated that the topological structure of the task space can be embedded as multiple attractors in the RNN by learning sensorimotor experiences.

Tani's subsequent research shifted towards understanding how higher-order cognitive functions are self-organized through the learning of continuous sensorimotor experiences, with minimal constraints imposed on neuro-dynamical systems and without introducing explicit mechanisms.
Specifically, Tani et al. proposed hierarchical RNN models, such as the RNN with parametric bias (RNNPB)~\citep{Tani2003} and multiple timescale RNN (MTRNN)~\citep{Yamashita2008}.
These models introduce a temporal hierarchy~\citep{Kiebel2008, Kiebel2009}, where the lower level exhibits a high temporal resolution, which may correspond to System 1, and the higher level exhibits a lower temporal resolution, which may correspond to System 2.

In the RNNPB~\citep{Tani2003}, the frequency of neural dynamics updates differs between levels.
The lower level operates at the sensorimotor level with rapid updates, whereas the higher level updates more slowly.
The higher level provides a vector called PB as a prior to the lower level, and the lower level generates sensorimotor predictions based on this PB from the higher level.
The resulting prediction error, calculated as the difference between predicted and actual feedback, propagates upward to the higher level, updating its internal model of the world.
This mutual interaction between the top-down predictive and bottom-up recognition processes can be understood as the RNN implementation of a predictive coding scheme~\citep{Rao1999}.
Based on this RNNPB scheme, several attempts have been made to account for cognitive processes, such as object manipulation~\citep{Ito2006a}, imitation~\citep{Ito2006a}, language acquisition~\citep{Sugita2005}, and social interaction~\citep{Chen2016}.

\begin{figure}[bt]
    \centering
    \includegraphics[width=1.0\textwidth]{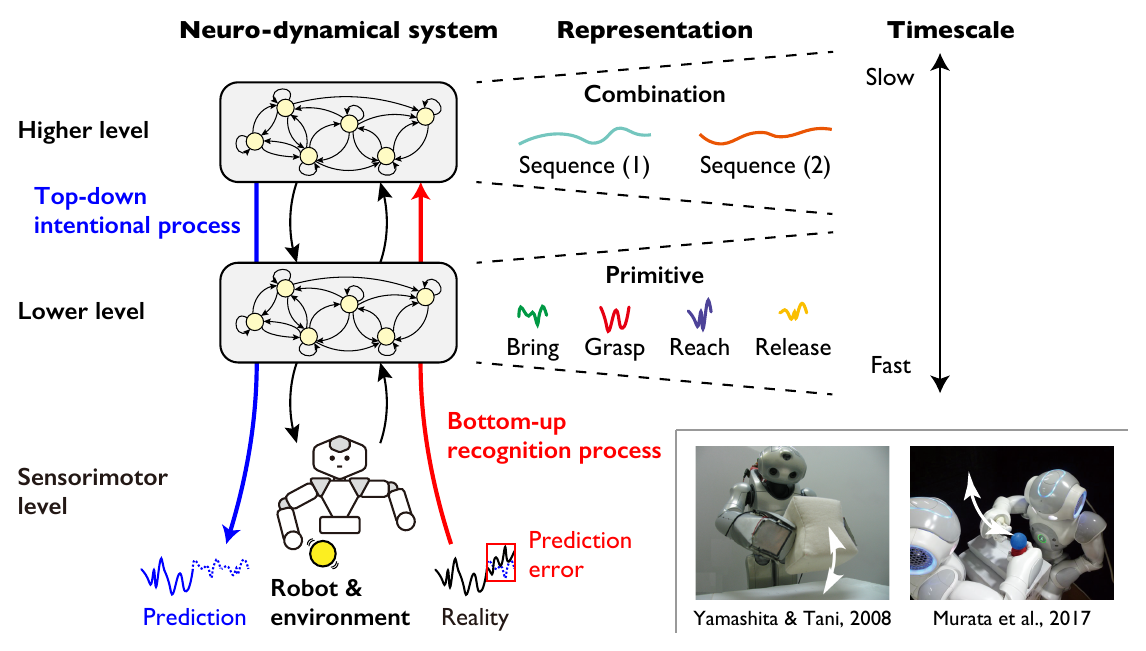}
    \caption{Neuro-dynamical system and examples of robotic tasks.}
    \label{fig:neuro_dynamical_systems}
\end{figure}

Unlike RNNPB, which features different frequencies of neural dynamics updates, the MTRNN incorporates multiple timescales by assigning different time constants to each network level~\citep{Yamashita2008}.
At a higher level, larger time constants produce slower neural dynamics, whereas at a lower level, smaller time constants lead to faster dynamics.
This temporal constraint on each level allows the MTRNN to self-organize the so-called functional hierarchy, where the lower level with fast dynamics captures detailed primitives of sensorimotor experiences, and the higher level with slow dynamics captures abstract combinatorial representations of the primitives (see Figure~\ref{fig:neuro_dynamical_systems}).
Several studies have demonstrated that primitives are acquired as limit cycles or fixed-point attractors at the lower level, whereas combinations of these primitives are represented as transient or chaotic dynamics at the higher level~\citep{Yamashita2008, Nishimoto2009, Namikawa2011, Murata2018}.
In addition to cognitive neurorobotics studies, MTRNN-based models have been applied to more practical situations, such as flexible object manipulation \citep{Suzuki2018} and door opening and entry with whole-body control~\citep{Ito2022}, demonstrating their engineering utility.

These fully deterministic neuro-dynamical systems, such as RNNPB and MTRNN, have been extended by incorporating uncertainty in both lower-level sensorimotor predictions (stochastic MTRNN; S-MTRNN)~\citep{Murata2015a} and higher-level latent representations (predictive coding-inspired variational RNN; PV-RNN)~\citep{Ahmadi2019}, as well as in the multiple timescale recurrent state-space model (MTRSSM)~\citep{Fujii2023}.
The formulation of these models aligns with the free-energy principle~\citep{Friston2005, friston2010free, Parr2022}.
These models demonstrate that noise at the lower level results in probabilistic transitions of primitives, whereas noise at the higher level produces similar phenomena.

Models such as MTRNN, S-MTRNN, and PV-RNN, with their multiple timescales and complex temporal representations, have been explored not only in cognitive neurorobotics but also in the context of computational psychiatry~\citep{Montague2012}.
Functional disconnections between hierarchies~\citep{Yamashita2012a, Idei2021a} or aberrant estimation of uncertainty~\citep{Idei2018, Idei2020} have been employed to explain pathologies in psychiatric disorders such as schizophrenia and autism spectrum disorder.
These models offer a promising direction for understanding how neuro-dynamical processes contribute to cognitive dysfunction.

\subsection{Embodied Systems with Morphological Computation (System 0)}\label{sec:system0}
Biological systems are physical but also intelligent systems, i.e., their own bodies, outside of their cognitive systems.
The system comprises pre-cognitive, embodied processes that occur at the super-fast timescale as our framework's foundational layer of cognition.
These processes emerge from direct physical interactions between an agent''s body and its environment without explicit computation, which is called morphological computation~\citep{pfeifer2001understanding}.
In other words, physical interactions generate the adaptive behavior of robots without computation using computers.
Then, the system ``computes'' any tasks using the physical system.
In addition, System~1 leverages System~0's computing results for recognition, control, and learning.

\begin{figure}[tb]
    \centering
    \includegraphics[width=0.99\textwidth]{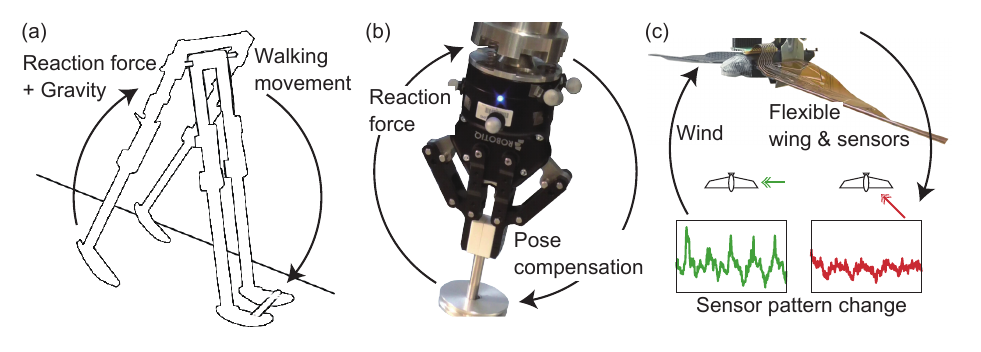}
    \caption{
    (a) Passive dynamic walker that walks without an actuator or computer.
    (b) Soft wrist that enables a robot arm to learn peg-insertion tasks using its passive dynamics.
    (c) Flapping-wing robot that computes the wind direction using the dynamics of its flexible wing movements.
    }
    \label{fig:system0}
\end{figure}

Morphological computation involves information processing by physical systems during the interaction between a robot's body and the environment.
The passive walker is a well-known example of morphological computation~\citep{mcgeer1990passive} (Figure~\ref{fig:system0}a).
This walker can walk down a slope without an actuator or computer.
The physical interaction between the walker's body, gravity, and slope generates the walking behavior. 
Another well-known example is a soft gripper~\citep{shintake2018soft}. 
This gripper can stably grasp objects of various shapes by passively changing its shape corresponding to the object and enclosing it with the same motor control signal.

Morphological computation using physical interaction has been used for System~1's control tasks of robots, as shown by the soft gripper.
% The physical interaction between the robot's body and the environment can be used as computational resources to control the robots, like the passive walker and the soft grippers.
Iida et al. reported that a legged robot with elastic elements could stably maintain hopping by using its dynamics~\citep{iida2004cheap}.
They showed that the elastic elements enable the hopping robot to maintain a stable attractor.
Hamaya et al. proposed a method for learning peg insertion using a robotic arm with a soft wrist~\citep{hamaya2020learning, von2020compact} (Figure~\ref{fig:system0}b).
They divided peg insertion into multiple sub-tasks and moved the arm using simple commands in each sub-task.
A robot using their method learns peg insertion efficiently owing to the simplicity of the commands in each sub-task and the passive deformation of the soft wrist corresponding to the contact state between the peg and hole.
Preflex is a faster response of the human muscles than neural reflex~\citep{dickinson2000animals}.
Human runners adjust the stiffness of their ankles by co-activating antagonistic muscles to modulate the reaction force against the ground.  

Theories of physical reservoir computing (PRC) describe System~0's computation using physical interaction~\citep{tanaka2019recent}.
PRC leverages the dynamical state changes of physical materials for computation, using sensor data as activation of nodes in RNNs.
Nakajima et al. reported that the complex movement of an octopus-like soft arm can be used as a computational resource through PRC~\citep{nakajima2014exploiting}.

Such computational results using physical interactions have been used for recognition.
Homberg et al. developed a method to identify an object that is grasped by soft fingers~\citep{homberg2015haptic}.
The soft fingers change their shapes according to the shape of the grasped object.
Therefore, the grasped object can be identified using the bending sensors on the fingers. 
Tanaka et al. reported that a flapping-wing aerial robot with soft wings and soft bending sensors on the wings could recognize the direction of the wind around the robot using PRC~\citep{tanaka2021flapping} (Figure~\ref{fig:system0}c).
In the flapping movements, the soft wings change their shape in a complex manner in response to the wind.
The bending patterns of the wings reflect the direction of the wind. 
Thus, the direction can be classified using a simple linear classifier based on the time-series data of the bending sensors.
Judd et al. proposed a method for estimating the location of an object in a water tank based on the movement of a soft arm in the tank using PRC~\citep{judd2019sensing}.
The arm stirs the water, and the location of the object changes the water flow in the tank, which changes the movement of the arm.
Therefore, the object's location can be estimated from the time-series data of multiple marker positions on the arm.

We refer to the layer mentioned in this subsection, which performs morphological computation, as ``System 0'', as defined in Section~\ref{sec:system0123}, within the framework of System 0/1/2/3.

\subsection{Multi-agent Systems with Emergent Language}\label{sec:system3}

Humans have demonstrated remarkable capabilities as a species, largely attributed to the invention and use of language as a tool for collaboration. Language is believed to have co-evolved with the human brain~\citep{deacon1998symbolic}, signifying a shift from individual to collective intelligence as the defining characteristic of our species. The idea of intelligence and life as collective phenomena has been a recurring topic of interest. The complexity and plasticity of language are fundamental to the formation of human societies and civilizations. Without these traits, the development of modern society and civilization would be difficult to imagine. In fields such as artificial intelligence, complex systems, and social decision-making, it is well established that collective intelligence often surpasses the capabilities of individual intelligence~\citep{surowiecki2004wisdom,miller2010smart}. In this study, we focused specifically on multi-agent systems (MASs) that utilize symbols and language.

LLMs have demonstrated how statistically modeling language can form the basis of intelligence. However, although much of the research on LLMs emphasizes improving the performance of individual models, there is a growing trend of investigating LLMs as components of MASs. Recent studies have highlighted the potential of using multiple LLM-based agents in MASs to explore collective behaviors enabled by language. For example, \citet{park2023generative} presented a groundbreaking study titled "Generative Agents," where they used computational software agents based on LLMs to simulate the lives of 25 agents in a small virtual town. This study inspired subsequent research on utilizing LLM agents as models of human behavior, enabling rich multi-agent interactions that model human societies. Research in this area is rapidly expanding~\citep{guo2024large}.

Research has also demonstrated that leveraging multiple LLM agents can improve the overall system performance. For instance, \citet{li2024more} showed that sampling and voting techniques, combined with increasing the number of agents, lead to enhanced LLM performance. Similarly, \citet{wan2024knowledge} proposed methods for knowledge fusion, where probabilistic distributions from multiple LLMs are aggregated to create a unified model. \citet{mavromatis2024pack} introduced a test-time fusion method, which uses perplexity-based approaches to integrate knowledge from multiple LLMs during generation.

Another promising direction involves fostering collaboration through debates or discussions among agents to improve their reasoning capabilities. \citet{chen2023reconcile} proposed RECONCILE, a multi-model, multi-agent framework designed as a roundtable discussion among diverse LLM agents, significantly enhancing their reasoning capabilities. \citet{motwani2024malt} introduced Multi-Agent LLM Training (MALT), a training approach where multiple agents collaborate in a debate setting to solve reasoning tasks. Inspired by neural scaling laws, \citet{qian2024scaling} explored how increasing the number of agents in multi-agent collaboration impacts performance. In addition, \citet{subramaniam2025multiagent} proposed the use of social interactions in multi-agent debates to fine-tune LLMs.

While these approaches optimize the inference capabilities of MASs by leveraging pre-trained language models, another critical research avenue is to understand how such languages and symbols originally emerge. This area of study, known as emergent communication (EmCom), emergent language (EmLang), or symbol emergence, explores the dynamics of how languages evolve and adapt over time through decentralized human interactions~\citep{peters2024survey,LazaridouB-2020-emergent,galke2025learning,steels2015talking,cangelosi1998emergence,taniguchi2024collective}. Language systems are inherently dynamic and change over time to describe a wide range of phenomena~\citep{deacon1998symbolic,peters2024survey,Steels97,Steels2007,wittgenstein2009philosophical,STEELS2011339,tomasello2005constructing,taniguchi2018symbol}. Importantly, language is a type of symbol, and one of the most crucial characteristics of symbols is their arbitrariness. We argue that the plasticity of language and symbol systems is a key source of adaptability in humans.

Throughout history, humans have adapted to their environments as collectives by transforming their symbolic systems, including their language. Building on discussions around EmCom, EmLang, and symbol emergence in robotics, \citet{taniguchi2024collective} introduced the collective predictive coding (CPC) hypothesis. This hypothesis reframes EmCom/EmLang from the perspective of generative cognitive systems. CPC posits that symbolic communication can be understood through generative modeling, both at the level of individual agents and across groups of agents. 
Based on the view of CPC, EmCom models have been developed as an extension of representation learning ~\citep{hagiwara2019symbol,taniguchi2022emergent,furukawa2022symbol,hagiwara2022multiagent,inukai2023recursive}. Beyond the emergence of external categorical signs and internal representations, ~\citet{hoang2024compositionality} explored the emergence of compositional word sequences within the framework of CPC, while \citet{you2024multimodal} and \citet{saito2024emergence} extended the framework to continuous signs, such as voice pitch and time-series data. 
For individual agents, the free energy principle (FEP) and active inference provide a generative framework for understanding cognitive processes. Extending this idea, CPC applies the principles of FEP and predictive coding (PC) at a societal scale~\citep{taniguchi2024cpcms}, conceptualizing the emergence of symbols and language as a decentralized Bayesian inference process of shared external latent representations.

CPC emphasizes that the dynamics of symbol emergence are intrinsically tied to the arbitrariness and adaptability of language. This hypothesis underscores the role of collective intelligence in symbolic systems, where language serves as a dynamic medium for communication and adaptation. Within the framework of System 0/1/2/3, as described in Section~\ref{sec:system0123}, we refer to this layer of language-based collective intelligence as "System 3." This conceptualization provides a foundation for understanding how language and symbols function as emergent properties of collective systems.

\subsection{Bergson's Philosophy: Multi-timescale interpretation}
The temporal hierarchy related to life and mind is an important topic within philosophy as well.
French philosopher Henri Bergson (1859–1941) is renowned for his distinctive theory of time. Recent studies, particularly from the Project Bergson in Japan (PBJ), have brought new insights into his work \citep{hirai2016,miyake_hirai_2018,tani2018,fujita2022,yoneda2022,hirai2022,hirai2023}. This new interdisciplinary approach is termed ``Expanded Bergsonism'' \citep{miquel2023}. According to \citet{hirai2022}, Bergson's philosophy of time differs from traditional approaches by emphasizing multiple intrinsic timescales. This subsection outlines Bergson's Multi-Timescale (MTS) interpretation in relation to System 0/1/2/3. The multi-timescale interpretation inspired by Bergson aligns naturally with contemporary cognitive architectures and robotics frameworks, particularly the hierarchical temporal dynamics observed in neuro-dynamical systems such as MTRNN and RNNPB, discussed in Section 2.1.2.

Bergson's concept of \emph{durée} (duration), initially introduced to describe human conscious experience of time, was expanded in his second major book \emph{Matter and Memory} \citep{bergson1896} to encompass the temporality of matter and various life forms from the perspectives of emergence and evolution. This concept of ``multiple rhythms of \emph{durées}'' can today be interpreted as a pluralism of timescales.

Traditional philosophical treatments of time have generally viewed it either a single cosmic absolute time or as subjective psychological time, often placing them in dualistic opposition. The notion of multiple intrinsic timescales has largely been overlooked. In contrast, Bergson focused on the proper timescales inherent to each local interaction system. His renowned example of ``sugar dissolving in water'' illustrates that each event has its own duration, ``which cannot be extended or shortened at will'' (\cite{bergson1907}, p. 10). Different solutes or solvents require different times. Bergson's philosophy resonates well with the concept of ``characteristic time'' in thermodynamics and condensed matter physics—concepts crucial to understanding system responses and relaxation processes.

Importantly, cognitive activities themselves are natural phenomena that possesses intrinsic timescales. Just as heat diffusion or chemical reactions require a specific durations, organisms also need specific intervals to perceive situations and perform actions. Thus, sensorymotor processes can be viewed as ``relaxation times'' required to re-stabilize the system. Bergson defined the ``thickness'' of the present as follows: the real present ``necessarily occupies a duration'' and is ``in its essence, sensorimotor'' (\cite{bergson1896}, p. 153). While he did not specify precise values, contemporary scholars such as Wilhelm Wundt and recent studies suggest the ``psychological present'' spans approximately 3-5 seconds \citep{poppel1985,fraisse1957,wittmann2016felt}. 

Within this thick present, temporal consciousness (\emph{durée}) resides. Changes that are too slow, such as celestial rotation or plant growth, are not directly experienced as ``time perception.'' Conversely, there is also a lower limit—the minimal temporal unit of perceptual elements. Bergson refers to ``temporal resolution,'' citing a value of around 2 millisecond (\cite{bergson1896}, p. 231, \cite{exner1894}).  Changes occurring more rapidly, such as those in fluorescent lighting, similarly escape our perception of time, resulting in phenomena like the wagon-wheel effect.

Thus, cognition operates within a specific temporal domain bounded by these upper and lower limits, termed here as the ``window of the present.'' Our consciousness perceives temporal changes exclusively within this domain. Bergson viewed this timescale-locality not as a limitation but as an  \emph{enabling} condition for consciousness itself. This particular ``slowness,'' relative to microscopic physical phenomena, is critical for the very emergence of consciousness~\citep{hirai2019,hirai2022}.

Moreover, as \citet{hirai2023b} clarified, Bergson also described a dual-process model of cognition based on the flexible nature of the window of the present: ``automatic recognition,'' which remains confined to the thin present, and ``attentive recognition,'' which leverages the thickness of the present, roughly corresponding to modern Systems 1 and 2 (Bergson, 1896, chap. 2). This point will be revisited in Section 4.2.

Bergson's philosophy of time, particularly with its emphasis on multiple intrinsic timescales, aligns with certain aspects of dual-process theories. Furthermore, it offers potential for broader discussions concerning the philosophy of consciousness and more expansive conceptions of time. Extending dual-process theory and elucidating its connections to Bergson's philosophy of time could prove crucial for advancing philosophical discourse in these areas.

\section{System 0/1/2/3: A Quad-Process Framework}\label{sec:system0123}

\subsection{System 0/1/2/3}\label{sec:system0123-subsec}
\begin{figure}[bt] % Example Figure
    \centering
    \includegraphics[width=1.0\textwidth]{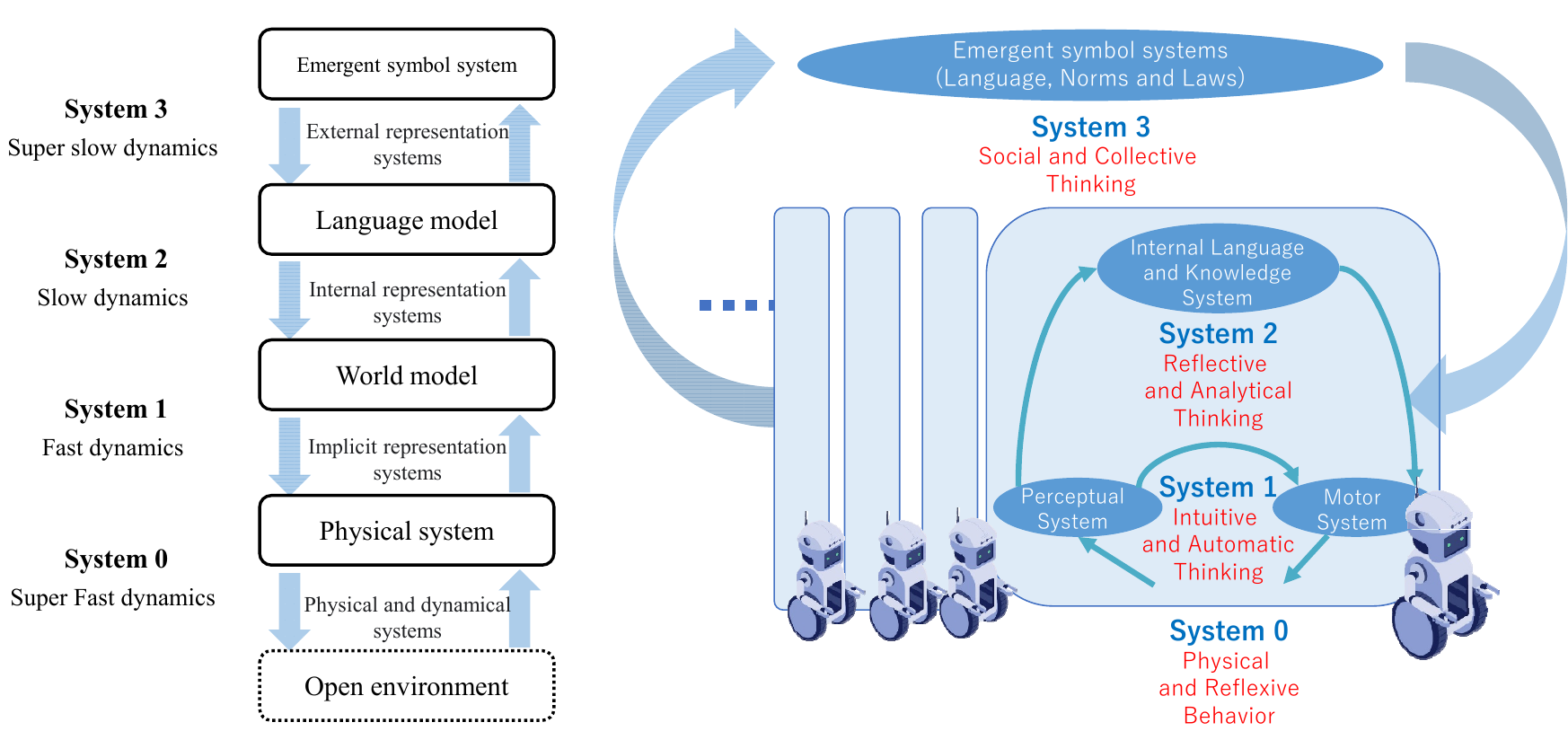}
    \caption{Hierarchical representation of System 0/1/2/3 cognitive framework}
    \label{fig:3-layer}
\end{figure}

This paper provides a novel view of the spatiotemporal hierarchy of human intelligence as a living system in the real world, generating symbols to perform semiotic communication and build societies and civilizations to thrive in this world. To this end, in this section, we introduce the framework of System 0/1/2/3. Figure~\ref{fig:3-layer} shows a schematic of System 0/1/2/3.

The dynamics of the body itself, which lie below System 1, perform morphological computation. 
System 0 corresponds to what is described in Section~\ref{sec:system0}. A well-designed body can realize adaptive behaviors such as walking, even without a brain, as proven by the passive dynamic walker.
System 0 represents the foundational layer of cognition in our proposed framework, encompassing the pre-cognitive, embodied processes that occur at the fastest (i.e., super-fast) timescale. 
The foundational idea behind System 0 can be philosophically grounded in Bergson’s notion of motor memory and canalization (introduced in Section~\ref{sec:Ber-Ex-Can}), highlighting how physical interactions pre-shape adaptive responses.
This system is characterized by direct physical interactions between an agent's body and its environment and operates without explicit computation. System 0 includes phenomena such as morphological computation and passive dynamics, exemplified by the self-stabilizing properties of bipedal walking or the inherent grasping capabilities of a soft robotic hand (see Section~\ref{sec:system0}).  These processes leverage the intrinsic physical properties of the body to perform adaptive behaviors and computations, often more efficiently than neural control alone. By incorporating System 0, our framework acknowledges the critical role of embodiment in cognition, highlighting how the body's structure and interaction with the environment contribute to intelligent behavior. This bottom-up approach to cognition provides a foundation for higher-level processes in Systems 1, 2, and 3 that influence perception, action, and even abstract reasoning by constraining and shaping the agent's interactions with the world.

For Systems 1 and 2, we follow the conventional definitions~\citep{kahneman2011thinking}. In connection with Bergson's philosophy, note that the dual temporal structure of cognition described in Section~\ref{sec:system12} aligns closely with Bergson’s dual-memory model—automatic recognition utilizing motor memory and attentive recognition leveraging pure memory (see Section~\ref{sec:Ber-Two-Type}).

The dynamics of semiotic, e.g., language-based, interactions in society lies above System 2. The uniqueness of human intelligence is in its adaptability that is not reduced to the individual. We create languages, norms, and social institutions to adapt to the environment as a group. Thus adaptation and creation process is not happening within our individual brains but operates in a distributed manner within the social group. This corresponds to symbol emergence and cultural evolution, including EmCom and EmLang. Humans not only adapt to the environment as individuals by constructing world models and creating internal representations but also create external representations, i.e., symbol systems (including language). 

System 3 represents the highest level of (super-)cognitive processing in our proposed framework, encompassing collective intelligence and symbol emergence. If the CPC hypothesis is correct, we as language creators engage in predictive coding or free energy minimization as an entire system~\citep{taniguchi2024collective}. 

This system operates on the longest timescale, facilitating the development and evolution of shared symbolic systems, such as language, cultural norms, and societal institutions. Unlike the individual-centric processes of Systems 1 and 2, System 3 emerges from the interactions between multiple agents within a social context, giving rise to distributed cognitive phenomena that transcend individual capabilities. System 3 extends Bergson's philosophical perspective on societal evolution and extended temporality, where collective symbols and cultural norms emerge over prolonged timescales (as discussed in Sections~\ref{sec:Ber-Three-Axes} and \ref{sec:Ber-Time-Can}).

System 3 plays a crucial role in human cognitive evolution and cultural development. It enables the accumulation and transmission of knowledge across generations, the formation of complex social structures, and the collaborative problem-solving capabilities that define human societies. Moreover, System 3 provides a feedback mechanism that influences the operations of Systems 1 and 2, thus shaping individual cognition through cultural learning and social norms. This bidirectional influence between individual and collective cognition highlights the deeply interconnected nature of the System 0/1/2/3 framework and underscores the importance of considering cognitive processes at multiple scales when developing artificial intelligence systems aimed at true human-like capabilities.

In recent years, the free energy principle (and active inference based on it) has been recognized as a model representing cognitive agents as individuals and has gained attention in neuroscience. If the dynamics of society as a whole can be understood similarly, we can view society itself —a collection of agents—as a subjective agent. This is System 3. In this sense, it may be more appropriate to call System 3 a super-cognitive rather than cognitive system.

\subsection{Collective Predictive Coding and System 3}\label{sec:cpc}

\begin{figure}[bt] % Example Figure
    \centering
    \includegraphics[width=0.8\textwidth]{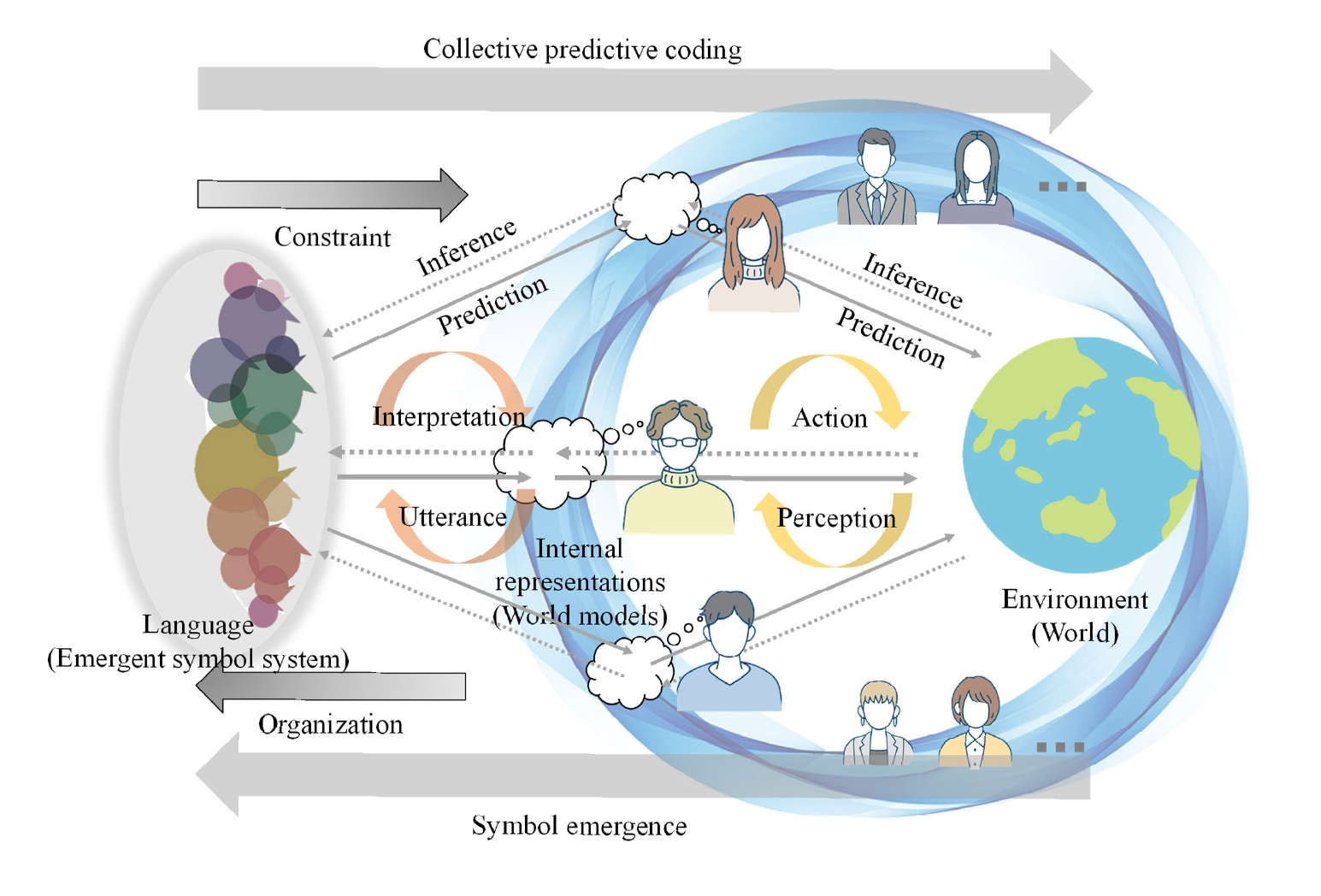}
    \caption{Overview of collective predictive coding hypothesis. \added{This figure is reused from ~\citet{taniguchi2024collective}}.}
    \label{fig:cpc}
\end{figure}

At the core of System 3 is the concept of collective predictive coding (CPC). In this system, groups of individuals collaboratively construct and refine shared mental models of the world through predictive processing and continuously update these models through social interactions and cultural transmission. This process leads to the emergence of symbolic systems that not only serve as efficient means of communication and coordination among group members but also as an external representation system modeling the world. 

As a basis for the view of System 0/1/2/3, we introduce the CPC hypothesis.
\citet{taniguchi2024collective} hypothesized that 
the emergence of language can be understood through the lens of CPC. 
A visual representation of CPC is presented in Figure~\ref{fig:cpc}. 

CPC expands the predictive coding concept~\citep{hohwy2013predictive} from individual cognition to collective societal systems. This framework allows the formalization of the emergence of symbol systems, including language, as a process of decentralized Bayesian inference~\citep{hagiwara2019symbol,taniguchi2022emergent}. Essentially, language, viewed as a distribution of sentences, is derived from the posterior distribution of latent variables, given the distributed observations of multiple agents.

The basic form of CPC introduced by \citet{taniguchi2024collective} can be mathematically expressed as
\begin{align}
\text{Generative Model:} \quad 
p(\{o_k\}_k, \{z_k\}_k, w) &= p(w) \prod_k p(o_k \mid z_k) p(z_k \mid w) \\
\text{Inference Model:} \quad 
q(w, \{z_k\}_k \mid \{o_k\}_k) &= q(w \mid \{z_k\}_k) \prod_k q(z_k \mid o_k)
\end{align}
where $w$, $z$, and $o$ represent symbols (including language), internal representations, and (perceptual) observations, respectively, following the conventions of the original and related papers.

\begin{align}
\text{Generative model:}\quad & p\left(w, \mathbf{z}, \mathbf{o} \mid \mathbf{a} \right) = p(w) p\left(\mathbf{o} \mid \mathbf{z}, \mathbf{a} \right) p\left(\mathbf{z} \mid w, \mathbf{a}\right) \\
\text{Inference model:}\quad & q\left(w, \mathbf{z}, \mathbf{o} \mid \mathbf{a} \right) = q\left(w \mid \mathbf{z}\right) q\left(\mathbf{o} \mid \right) q\left(\mathbf{z} \mid \mathbf{o}, \mathbf{a}\right) 
\end{align}
where $w$ represents external (collective) representations $\mathbf{z} = \{z^{k}\}_k$ is the set of representations by the $k$-th agent, $\mathbf{o} = \{o^{k}\}_k$ is the set of \added{sensory} observations by the $k$-th agent, $\mathbf{a} = \{a^{k}\}_k$ is the set of sequence of actions taken by the $k$-th agent.
\added{Here, the subscript $k$ indexes the agents in the group; for $K$ agents, $\{z^k\}_k$ is a shorthand for the set $\{z^1, z^2, \ldots, z^K\}$. We use set notation because the agents are an unordered collection in a decentralized system where no specific agent ordering is assumed.}
The inference of \( q\left(\mathbf{z} \mid \mathbf{o}, \mathbf{a}\right) \) corresponds to representation learning by the $k$-th agent, and \( q\left(w \mid \mathbf{z}\right) \) is assumed to be estimated through a language game within a group. Overall, symbol emergence by a group is performed to estimate \( q\left(w, \mathbf{z} \mid \mathbf{o}, \mathbf{a} \right) \) in a decentralized manner. We assume that we cannot estimate the true posterior $p\left(w, \mathbf{z}, \mathbf{o} \mid \mathbf{a} \right)$ but can only estimate its approximate distribution \added{$q\left(w, \mathbf{z}, \mathbf{o} \mid \mathbf{a} \right)$}. 

Based on these premises, the variational \added{collective} free energy $F$ of CPC, which corresponds exactly to the model's negative variational lower bound (ELBO), is defined as follows:
\begin{align}
F &= D_{\mathrm{KL}}\left[q\left(w, \mathbf{z}, \mathbf{o} \mid \mathbf{a} \right) \| p\left(w, \mathbf{z}, \mathbf{o} \mid \mathbf{a} \right) \right] \label{eq:cpc-fep}\\
&= \int q\left(w, \mathbf{z}, \mathbf{o} \mid \mathbf{a} \right) \ln{\frac{q\left(w, \mathbf{z}, \mathbf{o} \mid \mathbf{a} \right)}{p\left(w, \mathbf{z}, \mathbf{o} \mid \mathbf{a} \right)}} \mathrm{~d}w \mathrm{d}\mathbf{z} \mathrm{d}\mathbf{o} \\
&= \mathbb{E}_{q}\left[ \ln{\frac{q\left(w \mid \mathbf{z}\right)}{p(w)}} \right] + \mathbb{E}_{q}\left[ \ln{\frac{q\left(\mathbf{o} \right)}{p\left(\mathbf{o} \mid \mathbf{z}, \mathbf{a} \right)}}\right] + \mathbb{E}_{q}\left[ \ln{\frac{q\left(\mathbf{z} \mid \mathbf{o}, \mathbf{a}\right)}{p\left(\mathbf{z} \mid w, \mathbf{a}\right)}} \right] \\
&= \underbrace{D_{\mathrm{KL}}\left[ q(w \mid \{z^k\}_k) \| p(w) \right]}_{\text{Collective Regularization}}+\nonumber\\
&\ \ \sum_k\left[\underbrace{\mathbb{E}_{q}\left[-\ln{p\left(o^{k} \mid z^{k}, a^{k}\right)} \right]}_{\text{Individual prediction error}} +  \underbrace{D_{\mathrm{KL}}\left[ q(z^k \mid o^k, a^k) \| p(z^k \mid w, a^k) \right]}_{\text{Individual Regularization}} \right]
\label{eq:free_energy}
\end{align}

The inference of \( q\left(z^k \mid o^k\right) \) represents the representation learning process for the $k$-th agent. The inference of \( q\left(w \mid \{z^k\}_k \right) \) is achieved through linguistic interactions within a group, such as naming games~\citep{taniguchi2022emergent,hoang2024emergent,inukai2023recursive}. Collectively, these processes contribute to symbol emergence within a group, aiming to estimate \( q\left(w, \{z^k\}_k \mid \{o^k\}_k\right) \) in a decentralized manner.

\added{It is crucial to clarify that the FEP is assumed to operate hierarchically across all systems, starting from the sensorimotor level. The observations, o, and actions, a, in the CPC formulation represent precisely these sensorimotor signals. These signals are generated through the physical interactions of System 0 with the environment and are subsequently processed by System 1 and above, forming a continuous perception-action loop governed by the imperative to minimize free energy.}

Note that the symbols that emerge within System 3 are based on the arbitrariness and plasticity of symbol systems. The self-organization of System 3 is based on, so to speak, ``\emph{semiotic plasticity}.''
By contrast, Systems 1 and 2 are based on neuronal plasticity, which is realized by the collective regularization term in Equation~\ref{eq:free_energy}.

Now, we can consider MASs contributing to symbol emergence in society as an integrated cognitive process that minimizes free energy at the societal level. This symbol emergence is fundamentally based on embodied systems that interact with their environment. This view underpins the framework of System 0/1/2/3.

The subject of System 3 shifts the perspective from individual life to life as a collective entity. This can be understood by considering the question ``Who is minimizing the \added{collective} free energy described in Equations~\ref{eq:cpc-fep}-\ref{eq:free_energy}?'' The entity minimizing free energy is not \deleted{the}\added{any single} individual but rather the collective \added{group of} agents who share the variable $w$ defined in Equation~\ref{eq:cpc-fep},  representing a collective form of life that engages in predictive coding of the world. \added{This collective minimization, however, emerges from the decentralized behaviors of its constituents. As formulated in Equation \ref{eq:free_energy}, the collective free energy is composed of individual and collective terms. Each agent strives to minimize its own individual prediction error and regularization terms (the second and third terms in Eq. \ref{eq:free_energy}, respectively). At the same time, through inter-agent interactions such as the Metropolis-Hastings Naming Game~\citep{taniguchi2022emergent,hoang2024emergent,inukai2023recursive}, they collectively reduce the collective regularization term (the first term), which no single agent can minimize in isolation. In this way, the entire group descends the gradient of the collective free energy without a central controller.}

While System 3's operation corresponds to the dynamics of symbol and language emergence, one might assume that its learning relies solely on System 2 processes, such as individual language learning. However, we assert that System 3 is a collective emergent system. It adapts to its environment not only through the internal neural plasticity of individual agents but also by leveraging the plasticity of language itself in predictive coding. Consequently, the language formed through this process reflects collective predictive coding as a group.

\subsection{Relationships and Dynamics Among Systems: A Cognitive and Temporal Perspective}

\begin{table}[bt]
\centering
\scriptsize 
\caption{Relative Characteristics of System 0/1/2/3}	
\begin{tabular}{|p{2.8cm}|p{2.8cm}|p{2.8cm}|p{2.8cm}|p{2.8cm}|}
\hline 
\textbf{Characteristic} & \textbf{System 0} & \textbf{System 1} & \textbf{System 2} & \textbf{System 3} \\
\hline
Temporal Scale & Super Fast & Fast & Slow & Super Slow \\
\hline
Model & Physical and dynamical systems & World model & Language model & Emergent symbol system \\
\hline
Representation & Pre-representational & Implicit representations & Internal representations & External representations \\
\hline
Cognitive Process & Reflexive & Intuitive & Deliberative  & Collective \\
\hline
\shortstack{Consciousness \\\added{(Conscious Processing)}}& Non-conscious & Subconscious & Conscious & \added{(Supra-conscious?)} \\
\hline
Language Involvement & Non-linguistic & Sub-linguistic & \shortstack{Pre-linguistic\\(inner speech)} & \shortstack{Linguistic\\(outer speech)} \\
\hline
% Adaptability & Mechanistic & Reactive & Proactive & Evolutionary \\
% \hline
Scope of Interaction & Physical & Individual & Personal & Social \\
\hline
\end{tabular}
\label{tab:system-0123-comparison}
\end{table}

The framework of System 0/1/2/3 provides a hierarchical and structural distinction of each layer regarding cognition, representation, and temporal dynamics across varying levels of processes contributing to intelligent behaviors. Each system is characterized by unique temporal scales, cognitive processes, and interaction mechanisms, as outlined in Table~\ref{tab:system-0123-comparison}.

{\bf Temporal Scale and Model}: First, the systems operate across distinct temporal scales, ranging from the super-fast physical and dynamical processes of System 0 to the super-slow emergent symbol systems of System 3. System 1, associated with fast processes, incorporates intuitive decision-making, whereas System 2, operating at slower speeds, is widely accepted in dual-process theory~\citep{kahneman2011thinking}. From the viewpoint of recent AI studies, Systems 1 and 2 correspond to world and language models~\citep{taniguchi2023world,friston2021world}, respectively. 

{\bf Representation and Language}: The type of representation characterizes the nature of each system layer. System 0 automatically processes physical information without involving representations. System 1 handles implicit representations, such as non-linguistic features, which align with the concept of tacit knowledge~\citep{polanyi1966tacit}. System 2 employs explicit internal representations, corresponding to explicit knowledge. Although internal representations are not equivalent to language, they can be associated with inner speech, reflecting explicit internal processes. Finally, System 3 operates using external representations, corresponding to communication, such as outer speech between individuals.

When representing Systems 1 and 2 using a single neural network, long-term parameter changes correspond to learning, while short-term parameter adjustments are associated with inferences. By contrast, System 3 represents long-term parameter changes as symbol emergence and short-term parameter changes as communication. System 3 operates with slower dynamics than those of System 2. MASs leverage communication between agents during the reasoning process, aligning with the perspective of System 3, where communication corresponds to a collective reasoning process within collective intelligence.

\deleted{\textbf{Cognitive Processes and Consciousness}: Each layer is characterized by distinct cognitive processes and levels of consciousness. System 0 operates reflexively, without any form of consciousness, whereas System 1 engages in intuitive cognitive processes at the subconscious level. System 2 is associated with conscious reflection, whereas System 3 involves supra-conscious collective cognition. The relationship between cognition and language evolves across the systems, highlighting their roles in the emergence of consciousness. System 0 operates in a reflexive manner without consciousness, whereas System 1 engages in an intuitive cognitive process while being subconscious. System 2 represents a conscious deliberation. By contrast, System 3 undergoes supra-conscious collective cognition.}

\added{\textbf{Cognitive Processes and Conscious Processing}: Each layer is characterized by distinct cognitive processes, as summarized in Table 1. System 0 operates reflexively, without any form of consciousness. The distinction in the table regarding System 1 (``Subconscious'') and System 2 (``Conscious'') refers primarily to the concept of conscious information processing, common in dual-process theories. In this view, System 1 manages tasks automatically, without conscious deliberation, whereas System 2 involves explicit, conscious processing.}

\added{This discussion of information processing, however, must be distinguished from the broader issue of consciousness as subjective experience, which includes phenomena like qualia. In that regard, we do not claim that consciousness is exclusive to System 2. Rather, we propose that the fast, sensory-level processing of System 1 provides the essential raw material for phenomenal consciousness. This hierarchical view is well-aligned with Bergson’s philosophy of time. He distinguished between foundational ``sensory qualities'' (qualités sensibles), which is the equivalent of modern qualia, arising at the minimal temporal resolution of perception, and their integration into a thicker ``duration'' (durée) where meaningful, flowing experience is formed. Our framework maps onto this hierarchy: System 1 is associated with the emergence of these foundational sensory qualities, while System 2's reflective consciousness operates within this wider "duration," acting upon the materials provided by System 1.}

\added{System 3, in turn, invites speculation about collective cognitive phenomena. The notion of cognitive phenomena that transcend the individual has a long history in various schools of thought, exemplified by concepts like Durkheim's ``collective consciousness'' or Jung's ``collective unconscious''~\citep{durkheim1893division,jung1969archetypes}. Our framework, with its specific focus on temporality, allows us to pose a related but distinct question: could the vast temporal difference between individual cognition and the super-slow evolution of societal symbol systems give rise to a form of consciousness? While this remains a profound open question beyond the scope of this paper, we use the term "supra-conscious" to denote this hypothetical possibility—a collective cognitive phenomenon that is yet to be explored.}

{\bf Scope of Interaction}: System 0 pertains to physical interactions. By contrast, Systems 1 and 2 correspond to individual and personal interactions, respectively. Here, individual refers to basic and direct interactions between the individual and external environment, while personal involves deeper interactions characterized by internal processes, emotional connections, and meaningful relationships with others. System 3, in turn, corresponds to social interactions mediated by semiotic signs.

This hierarchical framework underscores the interplay and dynamics among the systems, highlighting how temporal scales, cognitive processes, and representational mechanisms converge to form an integrated perspective on cognition and interaction. By understanding these relationships, we can gain deeper insights into multi-scale spatial and temporal human intelligence as a living system with the abilities of symbol emergence and CPC.

\section{Extended Bergsonian MTS for System 0/1/2/3}\label{sec:mts}

Bergson's philosophy provides a grounding framework for the above-discussed extended four-tiered system in philosophical terms. This is because his philosophy of time encompasses not only the ``fast and slow'' processes of conscious cognition addressed in his second major work \emph{Matter and Memory}, but also issues of evolution and society, corresponding respectively to his third and fourth major works. In this section, we introduce Bergson's approach, which combines the ``landscape,'' an extended notion of space, with the ``window of the present,'' rendered elastic, as a way to structure and clarify the conceptual foundations behind such a wide range of phenomena.

\subsection{Three Axes of Timescales}\label{sec:Ber-Three-Axes}
\begin{figure}[tb]
\centering
\includegraphics[width=\textwidth]{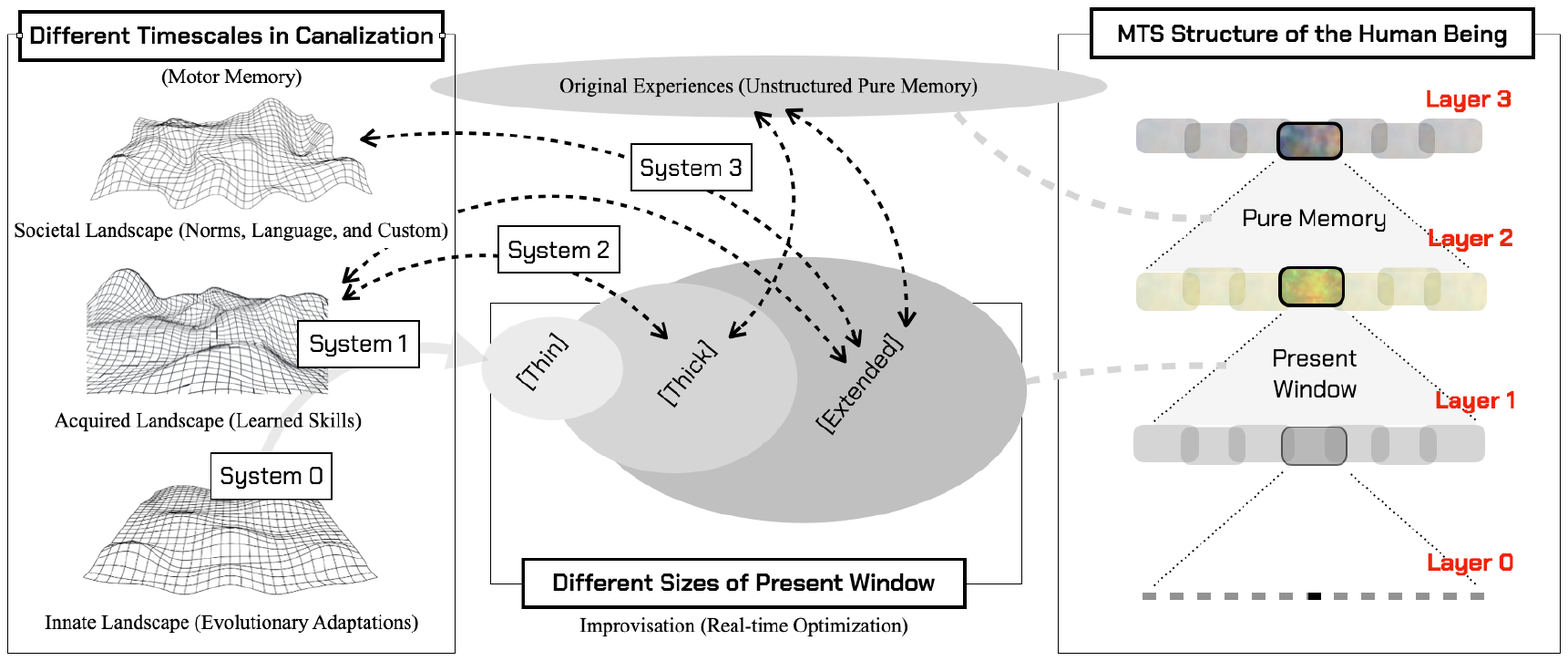}
\caption{Three Axes of Timescales in Bergson's Philosophy}
\label{fig:bergson}
\end{figure}

Bergson's philosophy of time offers a broad perspective on diverse phenomena, ranging from matter to society, each operating on different timescales. When distinguishing between different these timescales, \deleted{there are }it is beneficial to differentiate between multiple axes\deleted{ that}, which should not be conflated (see Figure~\ref{fig:bergson}).

The first axis refers to the multiple temporal layers that structure an individual organism. Centered around the present window—framed by layers 1 and 2 as described earlier—, it comprises a material layer below (layer 0) and a memory layer above (layer 3) (see the MTS structure: on the right side of Figure~\ref{fig:bergson}). The second axis represents the timescales required for the formation of adaptive optimization (``landscape'' formation or ``canalization,'' discussed later; see left of Figure~\ref{fig:bergson}). Evolution, for instance, involves timescales ranging from thousands to millions of years, while learning involves timescales ranging from weeks to months. The third axis represents the size of the present window (middle of Figure~\ref{fig:bergson}). A thicker present window enables real-time access to a richer reservoir of pure memory in layer 3. Strictly speaking, this is not a difference in time ``scales,'' but it serves as a decisive factor influencing higher-order emergent properties, such as deliberation and verbal ingenuity.

In real-world phenomena, these multiple axes of timescales intertwine and manifest together. Taking the systems proposed in this paper as examples: Both Systems 0 and 1 utilize only a ``thin'' window of the present, but the timescales of the underlying canalization differ between evolution and learning. System 2 has a ``thick'' present window, enabling access to memory, and interacts with the learning landscape. System 3 has an ``extended'' present window and additionally interacts with a collectively formed societal landscape.

\subsection{Two Types of Memory and Thickness of the Present}\label{sec:Ber-Two-Type}

Bergson's theory of time explains natural phenomena through the combined effect of two types of ``memory.'' This distinction is often seen as a precursor to modern concepts of procedural and episodic memory \citep{tulving2005,polster1991}; however, it has a broader scope. One type is ``motor memory,'' which refers to a series of motor patterns learned through repetition, such as skills and habits in humans. The other type is ``pure memory,'' which preserves original experiences \deleted{before they become organized into habitual patterns}\added{independently of their eventual consolidation into adaptive routines}. For example, through repeated trials of bicycle riding (r1, r2, r3), a motor memory for riding a bicycle (R) is acquired. However, one can also recall r1 as a specific life event on a particular date. Pure memory is what retains these unique instances.

Critically, the organization of motor memory, as an adaptation to the environment, operates under the constraints of utility at the given moment, whereas pure memory provides a temporally open resource—offering a surplus beyond these established habitual patterns. This surplus allows for flexibility and adaptability in novel situations that may arise later. Pure memory, with its flexibility and temporal openness, can be conceptually related to external memory systems utilized in recent machine learning approaches, such as Neural Turing Machines and Transformer architectures equipped with external memories.

Let us now revisit the two types of recognition briefly mentioned in Section 2.4: a fast but limited automatic process and a slower but intentional process, the latter termed ``attentive recognition'' (\cite{bergson1896}, chap. 2). Both automatic and attentive recognition rely on motor memory and share the function of facilitating interactions within stereotyped environments. However, automatic recognition, operating with a ``thin'' window of the present, is confined to immediate motor responses. In contrast, attentive recognition, with its ``thick'' window, provides access to pure memory. This temporal opening allows active reconstruction of the recognized object as a mental image, and in some cases, even enables re-experiencing scenes from one's personal past. Such temporal distinctions also likely apply to Systems 1 and 2.

Motor memory is essential for streamlining immediate cognition and action. Without this, complex motor responses would need to be assembled from scratch each time, making hierarchical and intricate actions unattainable. However, if all experiences were solely dedicated to constructing motor memory without leaving any surplus, there would be no additional resources to draw upon in novel situations.
The size of the present window acts as a constraint on this real-time improvisation. Thus, these dual memory resources are fundamentally important for intellectual pursuits and creativity. They enable preparedness and support through motor memory while retaining the freedom to deviate when necessary.

\added{Husserl, James, and Bergson—writing almost concurrently—converged on the idea that a bare, given succession or long-term memory cannot account for the flow experience of time; a distinct structure of time-consciousness is required. James and Bergson locate this structure in an extended but finite scope of the present, wide enough to encompass the lived flow of time, whereas Husserl’s retention–protention schema leaves that width formally unbounded; this classical contrast re-emerges today as the debate between \textit{retentional} and \textit{extensional} models of temporal experience \citep{Dainton2023}. Bergson goes further, treating the span of the present as an evolutionary variable: as it lengthens, organisms can integrate a broader sweep of ongoing sensory change within a single present, thereby enabling ever-richer forms of cognition \citep{Winkler2006}.}

\subsection{Extended Definition of Motor Memory: Canalized Landscape}\label{sec:Ber-Ex-Can}
In this paper, we define motor memory as self-organized ``prior optimization'' that occurs before actual experience (here, optimization does not necessarily aim for the global optimum). Mastering a movement and learning its patterns reduces cognitive and motor control demands when a similar situation arises again. Motor memory effectively serves as a form of \emph{temporal} outsourcing of optimization. This indicates that motor memory is not limited to internal mechanisms, such as updating motor programs or world models within an organism. Rather, organisms actively modify their environment through behaviors such as marking territory or building nests—and in humans, through artifacts like roads, buildings, and product design—to create conditions facilitating automatic responses that do not require conscious deliberation.

This definition, which focuses on the underlying temporal structure rather than the phenomenological classification, also captures adaptive optimization on an evolutionary scale, viewed as a large-scale self-organizing landscape. Such mechanisms lead to unconscious behaviors based on innate behavioral and morphological traits, corresponding to System 0 (Bergson calls this ``instinct'').

Bergson cites the example of the Sphex wasp stinging a caterpillar's nine nerve centers in a precise order to paralyze but not kill it (\cite{bergson1907}, pp. 174–175). This intricate maneuver is neither learned nor reasoned but pre-configured. On an evolutionary scale, this has been achieved through countless variations and selections. Thus, behavioral optimization is completed even prior to an organism's birth.

At both individual and evolutionary levels, establishing stable sensorimotor pathways in advance eliminates the need for a present window. In this sense, Bergson argues that as motor memory becomes perfected, it ``steps outside of time''(\cite{bergson1896}, p. 91). To illustrate this point further, we introduce the concept of the ``Canalized Landscape.''

Bergson's theory of perception involves the idea of knowledge outsourcing. In \emph{Matter and Memory}, he \deleted{rejects}\added{critiqued} both representationalism and atomistic associationism, proposing a form of ``direct realism'' grounded in an organism's potential actions \added{(\cite{Ansell-Pearson2018}, \cite{Guerlac2006})}. The idea that potential behavioral \added{``solicitations'' (\cite{bergson1896}, pp. 43-45) are distributed}\deleted{inducements are structured} within the environment not only anticipates von Uexküll's notion of the \emph{Umwelt} \citep{vonuexkull1909} but also aligns remarkably well with the dynamical systems perspective discussed in Section 2.1. Each organism operates within its own action-inducing space, which extends beyond its body and is shaped through motor memory processes, including evolution and learning. Bergson likened its formation process to river formation over geological timescales, calling it ``canalization'' (\cite{bergson1896}, p. 280, \cite{bergson1907}, pp. 94-95). Following C. H. Waddington, a renowned geneticist who inherited this idea via A. N. Whitehead \citep{posteraro2022}, we term the resultant space the ``landscape'' \citep{waddington1942}.

Thanks to this canalized landscape, organisms do not need to build behavior anew each time. Instead, they pre-construct an appropriate landscape in which the potential for appropriate actions is well arranged. Canalization shapes this potential space, and physical laws, evolutionary adaptations, and learning all contribute to it. Ideally, organisms achieve locally optimized behaviors by simply ``rolling down'' this landscape.

If a river meanders suitably, water does not need to ``think''; by flowing towards lower potential, it naturally follows the optimized ``valley'' paths. Similarly, the Sphex wasp's behavior arises not solely from internal motor control but also from coupling with the caterpillar's body shape and structure, allowing its actions to unfold automatically. Here, the spatial distinction between the internal and external becomes less significant. Instead, the focus is the \emph{temporally} outsourcing of landscape formation (canalization) prior to the window of the present\deleted{ window}.

\added{By slicing cognition along \textit{temporal} rather than spatial inside/outside lines, we obtain a staged and inclusive picture that neither depends wholly on representation nor banishes it. Treating representations as timescale–contingent strategies—emerging only when slower loops become relevant—avoids the rigidity of any single “-ism” and embodies the flexible stance (\cite{Constant2021}). This temporal–slicing scenario allows for a dynamic blending of enaction and representation in proportion to the available timescale: shorter loops favour direct sensorimotor engagement, whereas longer loops progressively introduce representational strategies. Bergson had already anticipated this logic: by focusing on how evolutionary elongation of an organism’s \textit{temporal architecture}—the delays built into sensorimotor coupling—creates room for ``memory–images,'' he explained the rise of attentive recognition and interpretive acts. In this sense, his view converges with the contemporary account of representations as timescale-dependent emergents (\cite{Bruineberg2014}).}

\subsection{Timescales of Canalization (System 0/1/2/3)} \label{sec:Ber-Time-Can}
The concept of canalization encompasses a variety of landscapes, ranging from physical laws to language, and naturally aligns with the hierarchical structure of Systems 0, 1, 2, and 3 proposed in this paper. Physical laws can be regarded as an extreme form of canalized landscapes. Physical matter, at least as understood in Bergson's era, does not make ``decisions''; its behavior is predetermined and immutable. This constitutes the physical landscape. Next, on the timescale of biological evolution, canalization shapes species-specific landscapes that correspond to their respective \emph{Umwelt}. Movements such as passive walking, which are automatically generated by morphological features (System 0), are positioned here. On the timescale of an individual's lifetime, learning further diversifies the folds of the landscape. The mastery of a skill leads to varying degrees of ease in performing actions and manipulating conceptual tools (Systems 1 and 2).

Finally, in humans who engage in social activities  through symbols, canalization also occurs on the timescale of cultural evolution. In his fourth and final major work, \emph{The Two Sources of Morality and Religion}, Bergson developed the concept of a socially canalized landscape \citep{Hirai2025}. When people act collectively, certain behavioral channels are carved into the social landscape to resolve coordination problems, such as social norms, rules, and morality. Similar to evolutionary and individual learning landscapes, societal canalization fosters community-specific repertoires of behavior. Without this level of canalization, language usage would not be possible (System 3; see Section~\ref{sec:system3}). 

Common to all these levels of landscapes is their dual nature: while they constrain possibilities by reducing choices, they also facilitate and promote stereotyped actions, reducing or even eliminating the need for the present window. Nevertheless, having a present window and access to unorganized pure memory holds significant importance for understanding the creativity inherent in life. 

First, Bergson reflects on the historical process by which the present window was gradually acquired through evolution. In simpler organisms, reactions ``can then hardly be delayed'' (\cite{bergson1911}, p. 22). Distant perception emerges when organisms become able to ``defer the date of their fulfillment'' (\cite{bergson1911}, p. 23) of responses. Thus, the presence of a present window and access to pure memory enables \emph{non-random improvisations} in unforeseen situations, facilitating complex adaptive behaviors beyond what evolution or individual learning alone can achieve.

Moreover, on-the-fly variability promotes learning efficiency. Bergson notes: ``thanks to this faculty, we have no need to await at the hands of chance the accidental repetition of the same situations, in order to organize into a habit concomitant movements; we make use of the fugitive image to construct a stable mechanism which takes its place'' (\cite{bergson1911}, p. 98). Similarly, machine learning exhibits improved efficiency when supplemented by working memory ~\citep{graves2016,vinyals2016}.

While the present window enriches phenomenological perception of time, its biological size is limited to just a few seconds. The invention of symbolic systems can be seen as artificially extending this limit, resulting in what we term the \emph{Extended Window of the Present} (EWP). In a later paper, Bergson wrote the following:

\begin{quote}
``Our consciousness tells us that when we speak of our present we are thinking of a certain interval of duration. What duration? It is impossible to fix it exactly, as it is something rather elusive. My present, at this moment, is the sentence I am pronouncing. But [$\ldots$] my attention is something that can be made longer or shorter, like the interval between the two points of a compass.'' (\cite{bergson1946}, pp. 178-179).
\end{quote}

Language stands out for its remarkable capacity to enhance the resolution and plasticity of human perception and memory, enabling a level of complexity unimaginable for organisms constrained by biological or personal limits. As Bergson observed in \emph{Creative Evolution} (1907), humans, unlike the socially organized yet instinct-driven hymenopterans, have reached an ``unlimited'' complexity. The present window can be extended through various means, such as conversation, symbolic note-taking, transactive memory, and incorporating cognitive resources of others during deliberation and dialogue. Such interactions with the social and collective landscape effectively serves as shared working memory, as illustrated by the CPC hypothesis (see Section~\ref{sec:cpc}). Thus, the inherently social nature of language is deeply rooted in the dynamic interplay between the landscape and the present window.

\section{Discussion}\label{sec:discussion}
The framework of System 0/1/2/3 and CPC allow us to tackle challenging topics involving social and physical dynamics, such as language, which should be grounded and cultural phenomena. This section discusses three important topics: AI, Alife, and robotics.

\subsection{Cultural Development\added{, Language Evolution,} and Creativity}
Language is a fundamental aspect of culture\added{, and research on language evolution} \deleted{Research on the evolution of language and culture} has often relied on the Iterated Learning Model (ILM). This model simulates the intergenerational transmission of language, modeling how parental language production and child language acquisition form a cycle. This explains how compositionality, a universal feature of human language, emerges through cultural evolution. This phenomenon has been demonstrated through agent-based simulations~\citep{kirby2001spontaneous,kirby2002learning,kirby2015compression}, mathematical models~\citep{brighton2002compositional,griffiths2007language,kirby2007innateness}, and laboratory-based language evolution experiments (experimental semiotics)~\citep{kirby2008cumulative,kirby2015compression,scott2010language}.
\deleted{The ILM was designed to capture the propagation of expressions.} However, it fails to adequately model the interaction between expressions and environmental adaptation, which involves physical interactions and (internal) representation learning, such as the generation of meaning for these expressions. 

By contrast, EmCom models focus on the process of forming external representations within a developmental framework~\citep{peters2024survey}. 
Building on this, Generative EmCom integrates generative model-based representation learning to establish a novel theoretical framework~\citep{taniguchi2024generative}. This framework, formalized as CPC, extends beyond language and encompasses a wide range of cultural expressions.
\added{Recent work has shown that the MHNG based on CPC can foster compositional languages, and that vision-language models based on the framework can realize emergent communication for complex visual tasks~\citep{hoang2024compositionality,matsui2025metropolis}. Furthermore, the CPC hypothesis is not merely theoretical; its core mechanism has received direct empirical support. An experimental semiotics study by~\citet{okumura2023metropolishastings} demonstrated that human participants' decisions to accept or reject a partner's proposed name are significantly better predicted by the acceptance probability calculated by the Metropolis-Hastings (MH) algorithm—a key component of CPC—than by other heuristic models \citep{okumura2023metropolishastings}.}

Artistic activities, such as music, literature, and performance, provide valuable avenues for exploration within CPC and the System 0/1/2/3 framework. Recent advancements in generative AI technologies, such as diffusion models for image creation and music generation techniques, have sparked debates regarding whether AI can truly possess creativity~\citep{ho2020denoising,rombach2022high,huang2024symbolic}. While much of this discourse focuses on the ``generation'' aspect, the essence of creativity extends beyond production to include ``evaluation'' and ``acceptance,'' as creative works are deeply rooted in human emotions and embodied sensory experiences. From a systemic perspective, creativity emerges through a dynamic interplay of various components, as emphasized by Csikszentmihalyi''s system model of creativity~\citep{csikszentmihalyi2015systems}. Creativity is not solely inherent in the artifact but arises from interactions between the creator, artifact, evaluating community, and contextual factors shaping its creation, such as cultural trends, historical events, and sociopolitical influences. These elements form a cyclical system in which each part continually affects and is affected by others; the artist draws inspiration from the community and context, while the community evolves its standards through the artist''s contributions.

Modeling such intricate dynamics within CPC and the System 0/1/2/3 framework poses a significant challenge for AI because it requires the integration of generative, evaluative, and emotional components to capture the reciprocal and emergent nature of human creativity.
Saunders et al. compiled a review of computational models of creativity that emphasize social interactions in the context of ALife. They also argued for the importance of studying creativity as a system theory~\citep{saunders2015computational}.
From a perspective of emotional feeling construction, discussions have emerged regarding the role of predictive coding in the sensation of interoceptions~\citep{seth2013interoceptive,barrett2015interoceptive}. By replacing exteroception, e.g., visuals and haptics, of sensory observations in CPC with sensations of interoceptions, artistic or entertaining expressions, including music, poetry, and performance, will fall within the scope of CPC, and therefore, within the scope of the System 0/1/2/3 framework. 
Considering the parallelism between music and language, the systematic correspondence between the emergence of two different symbol systems, i.e, music and language, has been discussed~\citep{taniguchi2021parallelism}.  

\added{Our full System 0/1/2/3 framework also leads to novel, falsifiable predictions for future research. First, it predicts that a group's shared symbol system will be systematically shaped by their physical embodiment (System 0). This could be tested in a robotics experiment comparing two groups of robots with different morphologies (e.g., a rigid gripper vs. a soft manipulator) to see if their emergent languages diverge to reflect their differing sensorimotor affordances. Second, the framework predicts that collective adaptation to a new environment will be more efficient through ``semiotic plasticity'' (re-negotiating the shared language in System 3) than through only individual learning (in Systems 1/2). This can be tested in agent-based simulations by comparing a group that can modify its language to one with a frozen language, measuring the speed of adaptation after an environmental shift.}

\subsection{Consciousness and its collectiveness}

As Bergson's discussions suggest, differences in timescales can be a fundamental source of consciousness. This perspective offers significant insights into the relationship between AI and cognition. Discussions on consciousness can be enriched by extending Bergson''s theory of timescales both downward and upward using the framework of System 0/1/2/3, thereby introducing new dimensions to the discourse.

Consciousness and unconsciousness also have a collective aspect, as exemplified by Jung's concept of collective unconsciousness ~\citep{jung1969archetypes}. The collective dimension of consciousness is a critical factor in understanding the human mind. Furthermore, Lacan highlighted the influence of language on the unconscious within the context of structuralism, emphasizing the role of linguistic structures in shaping the unconscious mind~\citep{lacan2006ecrits}.

Recent research has also explored the relationship between CPC and the structure of qualia~\citep{taniguchi2024constructive}. Building on System 0, the learning of representations through Systems 1 and 2 and the linguistic interactions facilitated by System 3 raise profound questions about how consciousness interacts with the structure of language formed in society, i.e., System 3. These interactions remain an open area of inquiry, particularly regarding how language and representation processes affect conscious and unconscious states.

Notably, the previously underexplored concept of collective unconsciousness could benefit from examination within the framework of System 0/1/2/3. This approach holds the potential to provide novel insights into consciousness, unconsciousness, and their collective aspects, with the aim of integrating these elements into a comprehensive understanding of the mind.

\subsection{Collective World Models and Robotics Foundation Models}
Recent advances in robotics have highlighted the development of robotics foundation models and vision-language-action (VLA) models, i.e., language-conditioned end-to-end robot controllers~\citep{arai2024covla,kim2024openvla,dey2024revla,zhen20243d,Kawaharazuka16092024}.
Explicit integration of Systems 1 and 2 into robotics foundation models and VLAs can be found in recent studies~\citep{zawalski2024robotic,chen2025fast,lin2025onetwovla,intelligence2025pi}.
Although these approaches build upon traditional methods of policy learning through neural networks and position-based control, they are gradually realizing the significance of integrating the influences of Systems 0 and 3.
System 0 introduces the importance of passive dynamics in robotics, while System 3 focuses on how language, shaped by social interactions, plays a critical role in robot control. These developments position robotics foundation models as a unifying research domain that bridges System 0/1/2/3. VLA models exemplify how language''s socially constructed properties can guide robotic behavior.

%From the perspective of CPC, the structure of language is inherently tied to the physical embodiment of the agent. Therefore, robots, which are artificial entities with bodies that are different from those of humans, may require different linguistic structures.
In CPC, an agent's physical form significantly influences its language structure. Consequently, robots—with physical forms distinct from humans—might develop unique linguistic structures tailored to their specific embodiments.
This highlights the need for EmLang models that account for symbol and language emergence in the context of robotics~\citep{taniguchi2016symbol}. Research in this area is becoming increasingly vital, as it seeks to integrate physical embodiment, social interaction, and cognitive dynamics into cohesive frameworks.

Additionally, System 3 emphasizes that language is not merely a medium for representing world models but serves as a collective world model itself~\citep{taniguchi2024generative}. Language encapsulates the shared experiences and knowledge of communities, functioning as a unifying representation of collective understanding. By adopting the perspective of System 0/1/2/3, these discussions gain clarity and provide a more comprehensive foundation for advancing robotics and AI. This integration opens pathways for innovative research on collective world models and their application in robotics.

\section{Conclusion}\label{sec:conclusion}

In this paper, we proposed the \emph{System 0/1/2/3 framework}, an extension of dual-process theory that integrates embodied, individual, and collective intelligence within a unified multi-timescale perspective. By introducing \emph{System 0} (pre-cognitive embodied processes) and \emph{System 3} (collective intelligence and symbol emergence), our framework captures the cognitive processes spanning \emph{super-fast} sensorimotor interactions to \emph{super-slow} societal-level adaptation. This approach bridges the existing gaps in cognitive science, AI, robotics, and artificial life, offering a more comprehensive perspective on intelligence that extends beyond traditional dual-process models. 

A key theoretical contribution of this work is the extension of \emph{Bergson's MTS interpretation} to account for the cognitive processes unfolding across diverse temporal hierarchies. By synthesizing insights from \emph{predictive processing, the free-energy principle, embodied cognition, and semiotic communication}, we provided a new lens through which to examine the interplay between neural, bodily, and collective intelligence. The integration of \emph{morphological computation and CPC} highlights the importance of both \emph{internal cognitive processes and external self-organizational dynamics}, paving the way for more holistic AI and robotics models.

Despite its strengths, our framework presents several open challenges and opportunities for future research. First, while we have provided a theoretical foundation, empirical validation is necessary to examine how \emph{System 0/1/2/3 dynamics} manifest in real-world embodied AI and human cognition. Future studies should explore experimental methodologies to quantify the influences of \emph{pre-cognitive embodied processes} (System 0) and \emph{collective intelligence} (System 3) in both artificial and natural systems. 

Second, the computational modeling of multiscale intelligence remains an open problem. Future AI architectures should explicitly incorporate hierarchical timescales by integrating \emph{fast-reactive, deliberative, and emergent collective processes}. The development of AI systems capable of evolving language, adapting symbol systems, and autonomously negotiating meaning in a dynamic environment is crucial for bridging the gap between LLMs and truly embodied intelligence.

Finally, our theory has significant implications for AI-human symbiosis and the long-term integration of AI into society. Understanding how intelligent agents, both biological and artificial, construct, modify, and negotiate meaning through semiotic interactions offers exciting prospects for AI alignment, social robotics, and human-AI collaboration. Further interdisciplinary research spanning \emph{cognitive science, AI, robotics, and artificial life} is necessary to refine and expand the System 0/1/2/3 framework.

By extending cognition beyond the traditional dual-process view, we hope that the proposed framework will serve as a stepping stone towards a more comprehensive understanding of intelligence—one that recognizes the intricate interplay between neural, bodily, and societal processes across multiple timescales. Intelligence is not confined to internal computation; rather, it emerges through dynamic interactions with the body, external environment, and broader social and economic systems in which it operates. By integrating these perspectives, our framework fosters interdisciplinary dialogue, linking cognitive science with fields such as semiotics, linguistics, humanities, and economics, where meaning-making, communication, and collective adaptation are fundamental. This broader perspective not only deepens our theoretical understanding of intelligence but also paves the way for more holistic approaches to AI, robotics, and human-machine collaboration.

\section*{Acknowledgment}
This work was supported by JSPS KAKENHI Grant Numbers JP21H04904, JP23H04835, JP23KJ0768, JP23H04974, JP23K28181, JP23K13304, JP24K03012, JP24K00005, JP23H04834,  JST PRESTO Grant Number JPMJPR22C6, JPMJPR22C9 and the JST Moonshot R\&D Grant Number JPMJMS2011.

\printbibliography
\end{document}